\documentclass{article}

\usepackage{arxiv}

\usepackage[utf8]{inputenc} 
\usepackage[T1]{fontenc}    
\usepackage{hyperref}       
\usepackage{url}            
\usepackage{booktabs}       
\usepackage{amsfonts}       
\usepackage{nicefrac}       
\usepackage{microtype}      
\usepackage{lipsum}
\usepackage{fancyhdr}       
\usepackage{graphicx}       
\graphicspath{{media/}}     
\usepackage{amsmath,amssymb,amsfonts}
\usepackage{bm}
\usepackage{orcidlink}

\title{Client Recruitment for Federated Learning in ICU Length of Stay Prediction
}

\author{
  Vincent Scheltjens\orcidlink{0000-0002-6382-5750} \\
  KU Leuven \\ 
  STADIUS Center for Dynamical Systems \\ Signal Processing and Data Analytics \\
   \And
  Lyse Naomi Wamba Momo\orcidlink{0000-0002-9019-0236} \\
  KU Leuven \\ 
  STADIUS Center for Dynamical Systems \\ Signal Processing and Data Analytics \\
  \AND
  Wouter Verbeke\orcidlink{0000-0002-8438-0535} \\
  KU Leuven \\
  LIRIS Research Centre for \\ Information Systems Engineering
  \And
  Bart De Moor\orcidlink{0000-0002-8438-0535} \\
  KU Leuven \\ 
  STADIUS Center for Dynamical Systems \\ Signal Processing and Data Analytics \\
}

\begin{document}
\maketitle

\begin{abstract}
\textbf{Background:} Machine and deep learning methods for medical and healthcare applications have shown significant progress and performance improvement in recent years. These methods require vast amounts of training data which are available in the medical sector, albeit decentralized. Medical institutions generate vast amounts of data for which sharing and centralizing remains a challenge as the result of data and privacy regulations. The federated learning technique is well-suited to tackle these challenges. However, federated learning comes with a new set of open problems related to communication overhead, efficient parameter aggregation, client selection strategies and more. 

\textbf{Methods:} In this work, we address the step prior to the initiation of a federated network for model training, client recruitment. By intelligently recruiting clients, communication overhead and overall cost of training can be reduced without sacrificing predictive performance. Client recruitment aims at pre-excluding potential clients from partaking in the federation based on a set of criteria indicative of their eventual contributions to the federation. In this work, we propose a client recruitment approach using only the output distribution and sample size at the client site. 

\textbf{Results:} We show how a subset of clients can be recruited without sacrificing model performance whilst, at the same time, significantly improving computation time. By applying the recruitment approach to the training of federated models for accurate patient Length of Stay prediction using data from 189 Intensive Care Units, we show how the models trained in federations made up from recruited clients significantly outperform federated models trained with the standard procedure in terms of predictive power and training time.
\end{abstract}


\section{Introduction}
\label{intro}
Recent machine learning and deep learning techniques have proven to be of significant value for healthcare applications \cite{mlhealth, mlhealth1, mlhealth2}. An abundance of medical data is continuously generated, especially in Intensive Care Units (ICU) where patients are monitored uninterrupted and vitals are charted continuously. Deep learning techniques are particularly good at extracting underlying complex relations in such large datasets, thus the vast amount of available data bodes well for the data hungry models. One of the main limitations in relation to medical data of any sort relates to data availability. The vast amounts of data are rarely aggregated on a central server where models can be trained. As a result of regulatory restrictions both in Europe through the General Data Protection Regulation (GDPR) \cite{european_commission_regulation_2016} and more recently the European Union Artificial Intelligence Act, and the United States with the Health Insurance Portability and Accountability Act (HIPAA) \cite{hipaa} imposing similar restrictions, data sharing poses a significant challenge in the medical sector. In both cases, sharing and centrally accumulating privacy-sensitive data is bounded by a legal framework, necessary to protect privacy of the individuals behind the data. This, however, poses a challenge for research and application oriented tasks that require large amounts of medical data. In practice, this often translates to significant delays as data sharing agreements need to be put in place and agreed upon by the stakeholders and their corresponding legal teams. 

In recent years, the value of federated learning (FL) has been illustrated for medical applications, especially in the field of computer vision for medical image segmentation and classification \cite{Rieke2020, ppbts, MIDL}. Here, the FL approach allows to learn complex models over decentralized data without direct access to the data. In doing so, there is no need to centrally gather data, which enhances privacy preservation. Note that there is no strict guarantee for privacy preservation as membership inference attacks remain possible on federated architectures \cite{FLattack}. The FL approach, however, comes with a new set of challenges. In the real world, decentralized data is often not independent and identically \textit{(non-IID)} distributed which harms the training procedure and model performance \cite{fliid}. The reduced performance can mainly be attributed to the weight divergence in local models as a result of the \textit{non-IID} data \cite{fliid}. 

Selection procedures of federated training algorithms may select those contributors whose data is most informative at training time. This, however, requires all potential contributors to form part of the federation and participate in at least one round of training. In \cite{Ruan}, the foundations are provided for a client recruitment process which precedes the initiation of the federation. This aims at building the federation with those clients for which it can, a priori, be said that their contributions to the federation will be valuable. More specifically, clients for which the local data does not form a good representation of the population are considered less representative and therefore, of less value to the federation. Intuitively, when aiming for accurate global predictive performance, less representative clients result in higher weight divergence in the local models at training time \cite{fliid}, which harms the training procedure and the predictive power of the resulting global model. 

Whilst in \cite{Ruan}, a sound optimization framework is provided for client recruitment, in this work, the central research objective is to investigate \textit{whether a subset of clients can be recruited by only looking at the local target distribution and sample size with the aim to train better performing federated models whilst reducing training time}. From the recruited subset of clients which are expected to yield better performing models naturally follows that overall training time is expected to reduce as well. 

\textbf{The contribution of this work is two-fold:} (i) building on \cite{Ruan}, we define a client recruitment approach and the construct of client-level representativeness using only the local output distribution and sample size; (ii) we demonstrate practical relevance by applying the proposed client recruitment approach to the real-world eICU dataset \cite{Pollard2018,  Pollard2017-iu, Goldberger2000-hm} containing data from 208 US hospitals for which we are able to learn better performing models to predict patient Length of Stay (LoS) in ICU with less clients in comparison to the FL approach without client recruitment. For this work, we consider the \textit{FedAvg} \cite{mcmahan17a} algorithm with random client selection as the standard FL approach and show how federated models for which only recruited clients partake in the federation significantly outperform the models trained using the standard approach in terms of predictive power and training time.

The remainder of this work is structured as follows. Section \ref{sec:related} outlines the related work followed by Section \ref{sec:fl} in which the basics of FL are discussed. In Section \ref{sec:methods} the client recruitment approach is discussed along with the experimental setup. Section \ref{sec:data} presents the data and Section \ref{sec:performance} outlines the corresponding results. In Section \ref{sec:conc} the findings are discussed and the work is concluded by discussing the corresponding limitations and directions for future work. 

The repository containing the code and instructions for reproducing the results is available on GitHub here: \href{https://github.com/vscheltjens/eicu-cl-recr}{github.com/vscheltjens/eicu-cl-recr}.

\section{Related Work}
\label{sec:related}
Federated learning was originally proposed by McMahan et al. in 2017 \cite{mcmahan17a} in conjunction with the \textit{FedAvg} algorithm and has since been proven to be a valuable learning framework that can yield accurate models without direct access to the local data. 

A significant amount of research efforts have been dedicated to the client selection problem. Client selection in FL deals with client scheduling for each round of training, i.e., for each training round a subset of clients is selected that will contribute in the next training iteration. The standard approach is to randomly select this subset. However, as argued in multiple studies \cite{clsel, clselectieee, clselesct}, a better approach is to impose criteria based upon which client selection can be performed. These criteria often relate to how informative updates from certain clients are, or the local computational resources \cite{ACL, MABCL}. This, however, does not allow for noninformative clients to be pre-excluded from the federation, which is where client recruitment, as extensively discussed in \cite{Ruan}, comes in. Client recruitment aims to discard potential clients from the set of available clients for the federation, prior to initiation of the latter. One of the ways to do so is by considering limited statistics on the local dataset that do not contain privacy sensitive information. By pre-excluding potential clients, the foremost benefit is that the overall cost of training a model in the federated setting is reduced \cite{Ruan} without sacrificing predictive performance. In addition, optimally, the less informative clients are pre-excluded which results in increased predictive performance for the resulting global model. 

Although FL originates in the large-scale distributed edge computing setting, it has been extensively studied in healthcare \cite{Rieke2020, ppbts, fleicu}. Specifically the work discussed in \cite{fleicu} closely relates to the work proposed here. In \cite{fleicu} the authors extensively assess different parameter settings for federated training on medical ICU data to identify suitable parameters for In Hospital Mortality (IHM) binary classification. In this study, we tackle a different problem, i.e., ICU LoS prediction, using the same eICU dataset. The authors show that a larger number of local training epochs improves performance whilst reducing the training cost at once as a result of the reduced server-client communication. In this work we extend upon these previous works by proposing a client recruitment scheme using limited statistics on the local data, applied to the critical care setting.

\section{Federated Learning Basics}
\label{sec:fl}

Federated Learning \cite{mcmahan17a} is a technique that allows for a central model to be trained over distributed data that originate and are stored locally. This approach was originally proposed within the setting of large-scale distributed learning on mobile devices and has been widely used and researched in the medical sector given the privacy enhancing aspect as well a the regulatory restrictions on data sharing within the sector. Each local data source, e.g., a mobile phone or a server hosted at a hospital, is referred to as a \textit{client}. The group of clients that contribute to training a central model, including the central \textit{server} is referred to as the \textit{federation}. The central server orchestrates the learning process over the different clients following the predefined FL algorithm. In that sense, the central server is holistically responsible for (i) initiating the model; (ii) providing a copy of the model to all of the clients; (iii) aggregating over the received model parameters and (iv) send back the updated model to the clients. Each client locally trains the model for a predetermined number of epochs after which only the model parameters are sent back to the central server. In the standard \textit{FedAvg} algorithm \cite{mcmahan17a} considered in this work, local model parameters are averaged into the global model update. In addition, for each communication round, either all or a subset of the clients in the federation are randomly selected for training. We consider this the standard FL setting which comes with most out of the box implementations.

In the original setting, where FL was envisioned for large-scale distributed learning on mobile devices, i.e., thousands of clients partake in a single federation, the server-client communication results in significant overhead depending on the training parameters. In this work, the clients correspond to 189 hospitals where communication overhead is not the main concern. The main challenge tackled here relates to some extent to the concept of \textit{client shift} in conjunction with the understanding that in the real-world, data from different hospitals is likely to be \textit{non-IID} distributed. More specifically, clients host data that originates from different hospitals in different geographic regions, each with potentially different demographic characteristics, etc. Depending on local sample size and training parameters, models may overfit to the local data and bias the global model when aggregating model parameters. These are common problems to FL training algorithms and have been tackled by introducing more involved aggregation algorithms such as \textit{Weighted FedAvg} or smart client selection \cite{clsel, clselectieee, clselesct}.

Whilst FL allows to learn a central model from distributed medical data in a privacy enhancing manner, not all data at each of the hospitals are equally valuable. More specifically, smaller local sample sizes are less likely to represent the global distribution which will typically result in a larger empirical training loss, as discussed in \cite{Ruan}. In addition, as a result of different local demographics, even larger local datasets could also diverge in distribution from the population, which will again reduce predictive performance. Existing client selection and aggregation algorithms provide partial solutions to these issues. These methods, however, require for all the clients to form part of the federation and for each client to have participated at least once in a training round. If not, the algorithms do not obtain the required information to guide the further aggregation and selection procedure. 

This work builds on the client recruitment work introduced in \cite{Ruan} and the field of FL in healthcare with the aim to develop a method for client recruitment by looking only at the local distribution of the target variable and the local sample size. With this, the recruitment process aims to recruit clients for which the local data better represents the population before initiating the federation. This would intuitively lead to better performing models globally and reduced training time. 

\section{Methods}
\label{sec:methods}

To assess the performance and utility of federated models that are trained with a subset of clients which are recruited following the approach described in \ref{ssec:clreq}, a single prediction task is defined for both the central model and federated models. As indicated, this work aims at verifying the proposed recruitment approach on real-world medical data. To that extent, the prediction task we define is for each model to predict the patient LoS in ICU, similar to what has been studied in \cite{rocheteau-eicu, Al-Dailami2022-rn, rocheteau, MMDL} where LoS is defined as the remaining time in ICU for a given patient.

Furthermore, we define (i) the model architecture which is used for both central and federated training in \ref{ssec:model} and (ii) the client recruitment approach in \ref{ssec:clreq} followed by a description of the central and federated training procedures in \ref{ssec:centraltrain} and \ref{ssec:fedtraining} as well as the metrics used to evaluate model performance in \ref{ssec:pe}. 

The prediction task for each of the models is to predict LoS using the first 24 hours of patient data post ICU admission. To that end, both temporal and static data are used denoted as $\bm{x}_{1:24}^i$ and $\bm{d}^i$ with $i$ the $i^{th}$ patient. The temporal data is fused with the static patient data which will serve as the input to the models. The formal prediction task is to yield predictions $\hat{y}$ which approximate $y$, the true LoS, equivalent to the difference between $T_D$ and $T_A$, the time at discharge and admission, respectively.

\subsection{Model architecture}
\label{ssec:model}

For both the centralized and federated training tasks in this work, the Gated Recurrent Unit (GRU) model architecture \cite{gru-paper} is used. As a subclass of the broader set of Recurrent Neural Networks (RNN), the GRU architecture is comprised of two sole gates. The reset and forget gates, respectively denoted as $r_t$ and $z_t$ in \eqref{eq:gru}. As a result of the dual gate architecture, GRU offers reduced computational complexity. This, in turn, is a desirable characteristic for FL applications where local computational resources and communication overhead pose real challenges \cite{fl-challenges}. 

In \eqref{eq:gru}, the governing equations for the GRU cell are shown. These outline the computations that occur for every discrete time step $t$ in the input sequence.

\begin{equation}
    \begin{split}
    & r_t = \sigma(W_{ir}x_t + b_{ir} + W_{hr}h_{(t-1)} + b_{hr})\\
    & z_t = \sigma(W_{iz}x_t + b_{iz} + W_{hz}h_{(t-1)} + b_{hz})\\
    & n_t = tanh(W_{in}x_t + b_{in} + r_t * (W_{hn}h_{(t-1)}+b_{hn}))\\
    & h_t = (1 - z_t) * n_t + z_t * h_{(t-1)},
    \end{split}
\label{eq:gru}
\end{equation}

with $x_t$ the input at timestep $t$, the latter of which corresponds to one hour in this work. In addition, $h_t, h_{(t-1)}$ denote the hidden states at time $t, t-1$ and $*$ represents the Hadamard product. 

The hidden state $h_t$, i.e., the output of the GRU cell, is provided as the input of a nonlinear Fully Connected Network (FCN), which yields a single output value representing the predicted LoS. The nonlinearity stems from the ReLU activation function leveraged in the FCN as shown in \eqref{eq:fcn}.

\begin{equation}
\label{eq:fcn}
    \hat{y}_t = ReLU\left(W_{y_t}h_t + b_{y_t}\right),
\end{equation}

with $\hat{y}_t$ the predicted value for LoS at time $t$. Employing $ReLU(x) = max(0, x)$ forces the outcome to be strictly positive. It is impossible for a patient to have a negative LoS, therefore we restrict the model to only yield predictions in the positive domain.

\subsection{Client Recruitment}
\label{ssec:clreq}
Consistent with the work discussed in \cite{Ruan}, we consider client recruitment to be a mechanism which is to be invoked prior to establishing a federation for training. To this extent, consider the set $S$ of $C$ potential clients with each a local dataset $D_{c} = \{(x_{i}, y_{i})\}_{i}$ where $x_{i}$ denotes the input data corresponding to the feature space and $y_{i}$ the target.

To facilitate client recruitment, we build on the work in \cite{Ruan} and let each potential client in $S$ report a tuple $(P_{co}, n_{c})$ to the central server where $P_{co}$ denotes the local distribution of the target and $n_{c}$ the local sample size. From here, the global sample size $n_{g}$ and output distribution $P_{go}$ can be calculated as shown in \eqref{eq:globals}.

\begin{equation}
\label{eq:globals}
    n_{g} = \sum_{c}n_c, \hspace{0.2cm}
    P_{go} = \sum_{c}P_{co}
\end{equation}

Using the tuple reported by the potential client, the local representativeness, similar to what has been done in \cite{Ruan}, of the client data $\nu_{c}$ in relation to the global dataset is calculated as a function of the output distribution divergence and the local sample size:

\begin{equation}
\label{eq:repr}
    \nu_{c} = \gamma_{dv}\bigg|\frac{P_{go}}{n_{g}} - \frac{P_{co}}{n_{c}}\bigg| + \gamma_{sa}n_{c}^{-0.5},
\end{equation}

where $\gamma_{dv}$ and $\gamma_{sa}$ denote weight parameters that influence the importance of the divergence of the output distribution and local sample size respectively. Furthermore, $\Tilde{P}_{c} - P_{c}$ converges to $\mathcal{N}(\mu,\,\sigma^{2})\,$ with $\mu = 0$ at the rate of $O(n_{c}^{-0.5})$, with $\Tilde{P}_{c}$ the empirical local distribution \cite{homem, SHAPIRO2003353}, from this follows that as $n_{c}$ grows larger, $\Tilde{P}_{c}$ better approximates $P_{c}$ \cite{Ruan}. Which is a desired characteristic for client recruitment. By inclusion of the term $n_{c}^{-0.5}$ in \eqref{eq:repr}, clients with larger local sample sizes are favored over those with smaller sample sizes.

The local distribution divergence in terms of the target denoted as $\big|\frac{P_{go}}{n_{g}} - \frac{P_{co}}{n_{c}}\big|$ is calculated as the difference between the normalized class counts locally and globally. In this work, the target corresponds to the patient LoS in fractional days. To facilitate the computation of the distribution divergence, ten bins are constructed with each bin corresponding to the number of target values (LoS in fractional days) that fall within that given bin. Conceptually, this is similar to class counts. The bins are defined as: $[(0, 1), [1, 2), [2, 3), ..., [7, 8), [8, 14), [14, +\infty)]$. This formulation converts the target from continuous to categorical for which the class counts are used to compute \eqref{eq:repr}. 

To select clients for recruitment, the per client representativeness values from \eqref{eq:repr} are sorted and stored in the vector $\bm{\nu}$. Consider:

\begin{equation}
\label{eq:globalrepr}
    \nu_{g} = \sum_{c}\nu_{c},
\end{equation}

where $\nu_{g}$ represents the global representativeness, which is used to define the recruitment threshold $\iota = \gamma_{th}\nu_{g}$ with $\gamma_{th}$ a configurable hyperparameter. Next, by summing over $\bm{\nu}$, the value $\nu_{c}$ at which the threshold $\iota$ is crossed, is identified. All the corresponding clients for values up until that point in $\bm{\nu}$ are recruited for the federation. This yields a subset of clients which are the most representative in terms of the target distribution divergence and local sample size. The obtained subset will be used for federated training, either considering all recruited clients or a fraction of the recruited clients in each training round, as will be discussed in \ref{sec:performance}. 

\subsection{Central training}
\label{ssec:centraltrain}
Central training is performed consistent with the traditional deep learning procedure in which all data is assumed to be centrally available. The architecture presented in \ref{ssec:model} is trained for a predetermined amount of epochs using the global train and validation sets, which consist of the accumulated data over all potential clients, i.e., originating hospitals. The loss is calculated as the Mean Squared Logarithmic Error (MSLE):

\begin{equation}
\label{eq:msle}
    MSLE = \frac{1}{n}\sum_{i=1}^{n}(\log(y_{i}+1) - \log(\hat{y}_{i}+1))^{2},
\end{equation}

with $y_{i}$ the true target value and $\hat{y}_{i}$ the predicted value. Model optimization is performed using \textit{AdamW} \cite{adamW}. Furthermore, the hyperparameters for training and optimization are fixed over all training iterations as shown in Table \ref{tab:gruhyps} with $L$ the number of layers, $N$ the hidden dimension for each of the layers, $\eta$ the learning rate, $m$ the batch size, $wd$ the weight decay for the \textit{AdamW} optimizer and $r$ the dropout. 

\begin{table}
    \centering
    \caption{Model hyperparameter settings used for both central and federated training}
    \begin{tabular}{l | c c c c c c}
        Model & $L$ & $N$ & $\eta$ & $m$ & $wd$ & $r$ \\
        \hline
        GRU  & 2 & 32 & 0.005 & 128 & 0.005 & 0.05 \\
        \hline
    \end{tabular}
    \label{tab:gruhyps}
\end{table}

\subsection{Federated training}
\label{ssec:fedtraining}
The FL procedure with and without client recruitment is simulated as a single process using the \textit{FedML} framework as proposed in \cite{chaoyanghe2020fedml}. More specifically, at training time, the central server instance is initiated along with the number of clients that form part of the federation. The central server instance initiates the model architecture as described in \ref{ssec:model} with the hyperparameters from Table \ref{tab:gruhyps} and distributes a copy to each of the clients in the federation. The clients each train the model on local train and validation data for a predetermined number of epochs. Upon completion, the clients report the learned model parameters back to the central server. This process is considered as a single round of server-client communication. Next, the server aggregates the received model parameters according to the aggregation algorithm. By aggregating over the received parameters, the central server constructs an update of the model and subsequently samples clients that will contribute in the next round of training. This latter process is referred to as client selection. In this work, clients are either all considered in each round of training or a subset is randomly sampled consistent with the standard implementation of \textit{FedAvg}.

For FL with client recruitment, the client recruitment process as described in \ref{ssec:clreq} is invoked prior to initiating the federation. Following the described implementation, the recruitment of clients is influenced by three user-defined hyperparameters $\gamma_{dv}, \gamma_{sa}$ and $\gamma_{th}$ which respectively define the importance of the divergence in the target distribution, the local sample size and fraction of the global representativeness to be covered by the recruited clients.

\subsection{Performance evaluation}
\label{ssec:pe}
For both central and federated training, the trained models are stored and subsequently evaluated against the hold-out test set containing unseen data from all hospitals in the data cohort. Meaning that for federated models for which clients were recruited prior to training, the test data contains data for patients residing in hospitals that did not contribute to training. 
Model performance is evaluated by running inference on the test set for which the following metrics are reported: (i) Mean Absolute Error; (ii) Mean Absolute Percentage Error; (iii) Mean Squared Error and (iv) the Mean Squared Logarithmic Error shown in \eqref{eq:metricspe} and \eqref{eq:msle}. In addition, training time\footnotemark is reported to evaluate time complexity for each of the models. \footnotetext{All models are trained on the same hardware. At training time, all models have 2 Xeon Gold 6140 CPUs@2.3 GHz (Skylake) with 18 cores each at their disposal and a total of 720GB RAM. All training was performed on the CPU architecture.}

\begin{equation}
    \begin{split}
        & MAE = \frac{1}{n}\sum_{i=1}^{n} |y_i - \hat{y_i}| \\
        & MAPE = \frac{1}{n}\sum_{i=1}^{n} \bigg|\frac{y_i - \hat{y_i}}{y_i}\bigg| \\
        & MSE = \frac{1}{n}\sum_{i=1}^{n} (y_i - \hat{y_i})^{2} \hspace{.2cm} \\
    \end{split}
\label{eq:metricspe}
\end{equation}

\section{Data}
\label{sec:data}

As indicated in \ref{sec:methods}, the client recruitment approach proposed in \ref{ssec:clreq} is evaluated by means of an application on real world-data corresponding to different institutions. To this end, the eICU dataset \cite{Pollard2018, Pollard2017-iu, Goldberger2000-hm} is used, covering over 200,000 eICU admissions to 208 US hospitals. The total admission count covers over 139,000 unique patients registered at one of the US hospitals between 2014 and 2015. 

The dataset is made publicly available for research purposes and is particularly interesting for FL as it allows for the data to be mapped to the originating institution. The data is preprocessed in concordance with the preprocessing pipeline proposed in \cite{rocheteau-eicu}. Due to limited space, we refer to \cite{rocheteau-eicu} for a detailed description of the pipeline. In summary, the first 24 hours of data post ICU admission for adult patients are extracted and used to predict LoS. Both temporal and static information are extracted, cleaned, re-sampled, imputed and one-hot encoded where needed. For patients with multiple recorded stays over the designated time period, only one of the stays is considered to avoid information leakage when obtaining train, test and validation splits. The summary statistics for the resulting data cohort are presented in Table \ref{tab:cohort}. 

\begin{figure}[htbp]
\centerline{\includegraphics[width=0.35\textwidth]{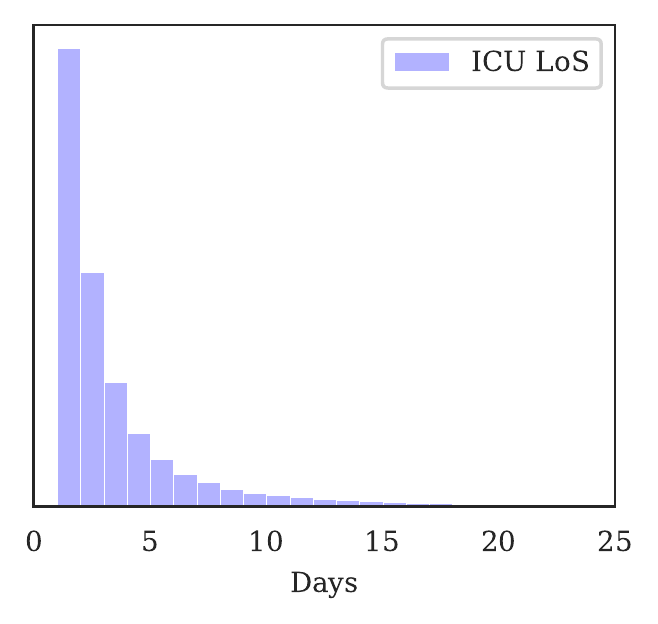}}
\caption{The distribution of the target (LoS) in the training data with a cutoff at 25 days on the x-axis. The y-axis represents the number of patients (frequency) with a certain LoS in days as indicated the x-axis.}
\label{fig:output}
\end{figure}

\begin{table}
    \centering
    \caption{Overview of the extracted and preprocessed data cohort}
    \begin{tabular}{ p{5cm} | p{1.5cm} }
        \hline
        Number of patient stays & 89,127 \\
        \quad Train & 62,375 \\
        \quad Validation & 13,376 \\
        \quad Test & 13,376 \\
        \hline
        Mean LoS & 3.69 \\
        Median LoS & 2.27 \\
        \hline
        Number of features & 38 \\
        \quad Temporal & 20 \\
        \quad Demographic & 18 \\
        \hline
        Number of hospitals (clients) & 189 \\
        \hline
    \end{tabular}
    \label{tab:cohort}
\end{table}

In addition, Fig. \ref{fig:output} depicts the output distribution for the training data. Essentially, when recruiting clients, the approach in \ref{ssec:clreq} aims at identifying clients that (i) have a reasonable sample size and (ii) best represent the output distribution as shown in Fig. \ref{fig:output}.

Furthermore, after preprocessing, 189 out of the original 208 hospitals are considered in the data cohort. In the federated training setting as described in \ref{ssec:fedtraining}, each of the clients correspond to one of the 189 hospitals. All clients contain local train and validation datasets which are subsets of the global train and validation dataset. The local data is a subset in the sense that a specific client only hosts the data for the originating hospital it represents. The test set is reserved to evaluate the performance of the resulting central and federated models.

\section{Results and Discussion}
\label{sec:performance}

At training time, the loss function and optimizer are fixed for both the central and federated training procedures. All training procedures leverage MSLE \eqref{eq:msle} as the loss function and \textit{AdamW} \cite{adamW} for optimization. For the central training procedure, the model is trained for 15 epochs, after which inference is ran against the hold-out test set. For the federated training procedure, four models are trained, each with specific settings in terms of the number of clients that partake in the federation and each training round and whether clients were recruited following the approach in \ref{ssec:clreq}. More specifically, the four models that are trained are:

\begin{itemize}
    \item \textbf{Federated-AC}: all clients make up the federation and partake in each training round.
    \item \textbf{Federated-SC}: all clients make up the federation, 10\% of which partake in each training round.
    \item \textbf{Federated-ARC}: recruited clients make up the federation and partake in each training round. 
    \item \textbf{Federated-SRC}: recruited clients make up the federation, 10\% of which partake in each training round.
\end{itemize}

When not all clients partake in each round of training, the selected subset is randomly sampled such that it represents 10\% of the clients that make up the federation. A summary of the settings specific to each of the models in the federated setting is provided in Table \ref{tab:fedparams}, with $\epsilon$ the total number of clients in the federation, $\delta$ the number of clients randomly sampled from $\epsilon$ in each round of training and ($\gamma_{dv}$, $\gamma_{sa}$, $\gamma_{th}$) the hyperparameters for client recruitment as described in \ref{ssec:clreq}. In addition, each client trains for four epochs per round of server-client communication for a total of 15 rounds. The resulting model is subsequently evaluated against the hold-out test set, the results of which are reported in Table \ref{tab:results} with $\tau$ the training time in seconds.

\begin{table}
    \centering
    \caption{Settings specific to the federated models}
    \begin{tabular}{l | c c c c c c}
        Model & $\epsilon$ & $\delta$ & $\gamma_{dv}$ & $\gamma_{sa}$ & $\gamma_{th}$ \\
        \hline
        Federated-AC  & 189 & 189 & - & - & -\\
        Federated-SC  & 189 & 19  & - & - & -\\
        Federated-ARC & 54  & 54  & 0.5 & 0.5 & 0.1\\
        Federated-SRC & 54 & 5  & 0.5 & 0.5 & 0.1\\
        \hline
    \end{tabular}
    \label{tab:fedparams}
\end{table}

\subsection{Client recruitment results}
\label{ssec:clrecres}

The results in \ref{tab:results} show the best performing models per metric in bold. When considering MSLE, which was used as the loss function for all models, the best performance amongst the federated models is obtained by Federated-AC. Here, all clients make up the federation and partake in each round of training. Note that the performance of Federated-AC is nearly identical to the central model in terms of MSLE. However, considering all clients at each round of training incurs significant overhead and slows down training, as illustrated by the high training time for Federated-AC. The second best performing model in terms of MSLE is Federated-SRC, trained with recruited clients, for which a subset was randomly sampled at each round of training. More specifically, from Table \ref{tab:fedparams}, we note that 54 clients were recruited from which 5 were randomly sampled at each round of training. This model outperforms Federated-SC and Federated-ARC in terms of MAE, MSE, MSLE and training time, whilst being the best overall performer in terms of MAE and training time. 

\begin{table*}
    \centering
    \caption{Model performance for central and federated models with and without client recruitment. Statistical significance among the federated models in comparison to the standard FL approach (Federated-SC) is indicated as $^{*}$ at the 5\% significance level and $^{**}$ at the 1\% significance level. The best results per metric for the federated models are shown in bold.}
    \begin{tabular}{*{15}{l | l l l l l}}
        Model & MAE & MAPE & MSE & MSLE & $\tau$(s)\\
        \hline
        Central & 2.21 $\pm$ 0.02 & 0.57 $\pm$ 0.06 & 21.94 $\pm$ 0.63 & 0.33 $\pm$ 0.01 & 2128 $\pm$ 18\\
        \hline
        Federated-AC & 2.26 $\pm$ 0.06 & 0.63 $\pm$ 0.08$^{**}$ & \textbf{21.61 $\pm$ 0.73$^{**}$} & \textbf{0.33 $\pm$ 0.02$^{**}$} & 5231 $\pm$ 29\\
        Federated-SC & 2.26 $\pm$ 0.06 & 0.46 $\pm$ 0.06 & 23.98 $\pm$ 1.26 & 0.41 $\pm$ 0.05 & 1469 $\pm$ 35\\
        Federated-ARC & 2.27 $\pm$ 0.12 & 0.57 $\pm$ 0.17$^{*}$ & 22.67 $\pm$ 1.83$^{**}$ & 0.37 $\pm$ 0.05$^{*}$ & 3359 $\pm$ 25\\
        Federated-SRC & \textbf{2.21 $\pm$ 0.03$^{**}$} & \textbf{0.46 $\pm$ 0.03} & 23.49 $\pm$ 0.73 & 0.37 $\pm$ 0.03$^{*}$ & \textbf{965 $\pm$ 24}\\
        \hline
    \end{tabular}
    \label{tab:results}
\end{table*}

The improved training time is a direct result of the reduced number of clients recruited for the federation. The improved performance can be attributed to the fact that model training is subject to less noise in the data, again as a direct result of the client recruitment procedure. Therefore, more informative model updates are produced at each round of training, resulting in lower empirical loss, which in turn improves performance. 

In summary, the client recruitment approach allows for models to be trained that outperform the standard FL approach (e.g., Federated-SC) and perform on par or better than the centrally trained model depending on the metric of interest. In addition, the total training time is drastically reduced compared to either central or federated models without client recruitment. In this use case, the model with the lowest MAE is of most value to the ICU, which is obtained by the Federated-SRC model in a fraction of the required training time compared to the other models in the experiment. Thus, illustrating practical relevance of the client recruitment procedure for larger federations in a real-world, privacy-sensitive, setting. 

\subsection{Recruitment parameter effects}
To gain additional insight in the behaviour of the client recruitment procedure proposed in \ref{ssec:clreq}, this section assesses the performance under different settings for the user defined hyperparameters $\gamma_{dv}$ and $\gamma_{sa}$. $\gamma_{dv}$ influences the importance of the divergence in the distribution between the local target distribtution and that of the target in the global data whereas $\gamma_{sa}$ affects the importance of the local sample size in the client recruitment approach. The parameters are set in such a manner that distribution divergence is prioritized over the local sample size and vice versa as follows: 

\begin{itemize}
    \item \textbf{Federated-SRC-QG}: Divergence over sample size with $\gamma_{dv} = 1$ and $\gamma_{sa} = 0.01$.
    \item \textbf{Federated-SRC-DG}: Sample size over divergence with $\gamma_{dv} = 0.01$ and $\gamma_{sa} = 1$.
\end{itemize}

This formulation is intuitively equivalent to the recruitment process being quality greedy (QG) in the former and quantity (data) greedy (DG) in the latter. The quality greedy recruitment strategy allows for clients with smaller sample sizes for which the output does not diverge significantly in distribution to be recruited. The contrary is true for the quantity greedy strategy in which the recruitment process neglects, to some extent, the distribution divergence in the output and favorably ranks clients with large local samples. Essentially, this approach explores performance in the extremes of the weighted function presented in \eqref{eq:repr}. 
The results in Table \ref{tab:parameterresults} show how nor the quality greedy approach, nor the quantity greedy approach perform better than the Federated-SRC model shown in Table \ref{tab:results}. In addition, we note how the data greedy approach results in increased training time due to the larger sample sizes in the local data of the recruited clients. The findings reported in Table \ref{tab:parameterresults}, when compared to the results for Federated-SRC in Tab. \ref{tab:results}, show the value of combining both divergence and local sample size in the client recruitment process as neither of the extreme strategies outperform the combined approach. 

\begin{table}
    \centering
    \caption{Model performance for federated training with data greedy and quality greedy recruitment strategies}
    \begin{tabular}{l | c c c c c}
        Model & MAE & MSLE & $\tau$(s)\\
        \hline
        Federated-SRC-QG & 2.23 $\pm$ 0.03 & 0.40 $\pm$ 0.03 & 891 $\pm$ 25\\
        Federated-SRC-DG & 2.22 $\pm$ 0.06 & 0.39 $\pm$ 0.05 & 1137 $\pm$ 15\\
        \hline
    \end{tabular}
    \label{tab:parameterresults}
\end{table}

\begin{figure}
\centerline{\includegraphics[width=0.8\columnwidth]{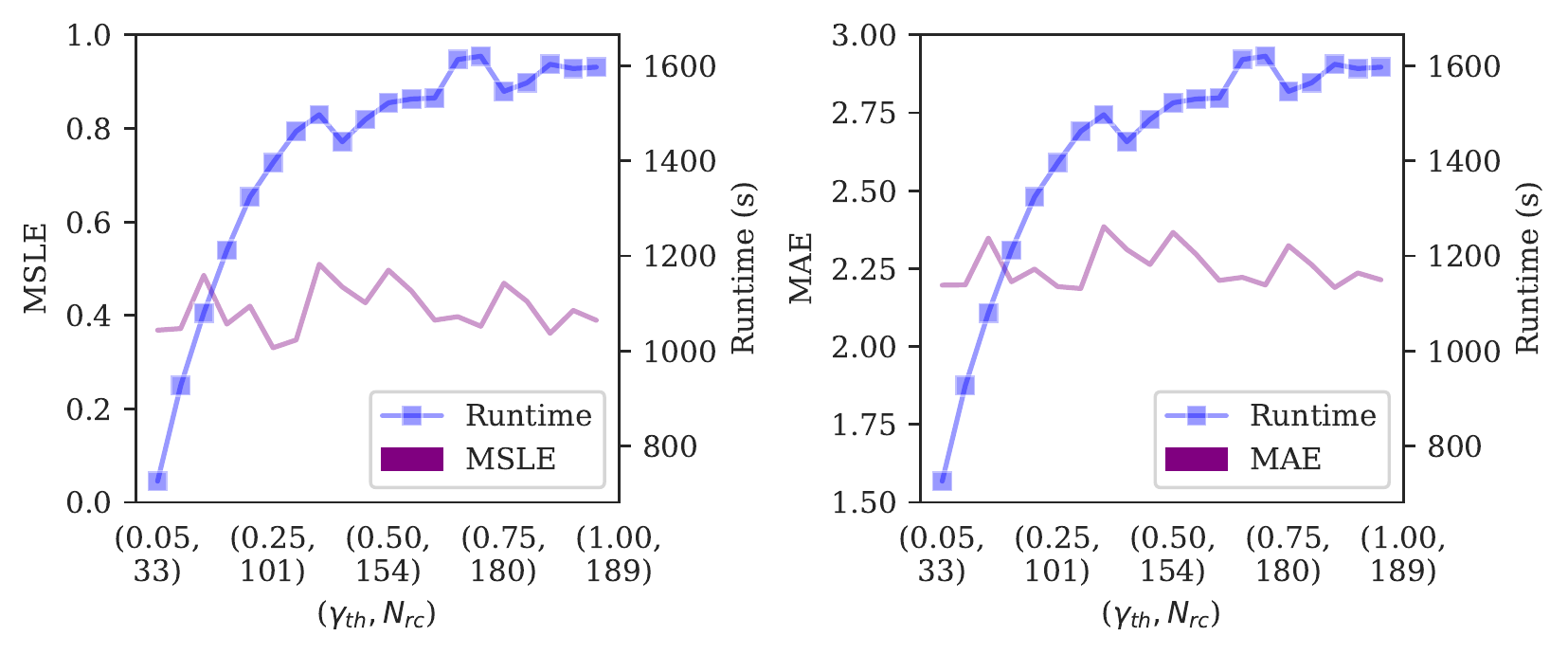}}
\caption{Runtime versus MSLE (left) and runtime versus MAE (right) in function of gradually increasing values for $\gamma_{th}$. Corresponding to a gradually increasing number of recruited clients, denoted as $N_{rc}$, for each of the training iterations.}
\label{fig:perfgamma}
\end{figure}

Having covered the effects on performance under different settings for $\gamma_{dv}$ and $\gamma_{sa}$, we investigate how different settings for $\gamma_{th}$ affect the performance and training time. Higher values for $\gamma_{th}$ directly correspond to more recruited clients for the federation, which in turn corresponds to higher training time. To this extent, the value for $\gamma_{th}$ is gradually increased in steps of 0.05. For each step we observe the performance on the test set in terms of MSLE, MAE and training time as shown in Fig. \ref{fig:perfgamma}. For $\gamma_{th} = 0.05$, 33 clients are recruited (denoted as $N_{rc}$ in Fig. \ref{fig:perfgamma}) and for each step, gradually more clients are considered until all clients make up the federation for $\gamma_{th} = 1$. Fig. \ref{fig:perfgamma} shows how there is no direct relation between performance and a higher number of recruited clients, even more so, close to optimal performance can be obtained for low values of $\gamma_{th}$, i.e., with few of the most representative clients making up the federation. 



\section{Conclusion}
\label{sec:conc}
\label{ssec:subconc}
In this work, we present a client recruitment approach considering only the local output distribution and local sample size. In addition, we show practical relevance of the proposed method in the medical setting. By recruiting clients in function of the herein defined client-level representativeness, those clients with smaller sample size in combination with those for which the output distribution vastly diverges compared to that of the global data are pre-excluded from the set of potential clients for the federation. By applying client recruitment, the predictive performance of the federated models significantly increases compared to the models trained with the standard FL approach. In addition, training time was significantly reduced as a direct result of the reduced number of clients that partook in training. 

This work builds upon existing work in the FL domain applied to healthcare and client recruitment research. The approach and results reported in this work further add to the foundation for the relevance and importance of intelligent client recruitment strategies as this can be a driver for overall performance of the federated training procedure. 

\section{Limitations and Future Work}
\label{ssec:limits}
The main limitation of this work stems from the introduction of the recruitment parameter, $\gamma_{th}$, which directly affects the number of clients recruited for the federation. In a real-world setting, tuning this parameter is not always feasible. In addition, the present work is executed in a simulated, single process environment, which does not affect the performance of the client recruitment scheme. However, this does mean that the improvement in training time is relative and the actual communication cost is not a factor. 

Future work will look at how to, a priori, approximate the optimal setting for $\gamma_{th}$. In addition, future work will evaluate performance in a real-world setting were data is hosted on actual servers corresponding to the hospitals in separate networks. Here, client recruitment is of even greater importance as it can greatly reduce the overall required server-client communication. Furthermore, future work will explore alternative recruitment strategies with a focus on sampling from diverse subgroups in the data and assess local performance of the federated models against models trained on the local data only. At last, in future work we will evaluate the performance of the herein proposed client recruitment strategy against the less widely adopted state-of-the-art aggregation and selection strategies in FL.



\section*{Acknowledgments}
This work was supported by KU Leuven: Research Fund (projects C16/15/059, C3/19/053, C24/18/022, C3/20/117, C3I-21-00316), Industrial Research Fund (Fellowships 13-0260, IOFm/16/004, IOFm/20/002) and several Leuven Research and Development bilateral industrial projects; Flemish Government Agencies: FWO: EOS Project no G0F6718N (SeLMA), SBO project S005319N, Infrastructure project I013218N, TBM Project T001919N; PhD Grants (SB/1SA1319N, SB/1S93918, SB/1S1319N), EWI: the Flanders AI Research Program VLAIO: CSBO (HBC.2021.0076) Baekeland PhD (HBC.20192204) and Innovation mandate (HBC.2019.2209) European Commission: European Research Council under the European Union’s Horizon 2020 research and innovation programme (ERC Adv. Grant grant agreement No 885682); Other funding: Foundation ‘Kom op tegen Kanker’, CM (Christelijke Mutualiteit)


\bibliographystyle{unsrt}  
\bibliography{references}

\end{document}